\documentclass[10pt,twocolumn,letterpaper]{article}

\usepackage{wacv}              
\usepackage{multirow}
\usepackage{microtype}

\definecolor{wacvblue}{rgb}{0.21,0.49,0.74}
\usepackage[pagebackref,breaklinks,colorlinks,allcolors=wacvblue]{hyperref}

\def\wacvPaperID{53} 
\def\confName{WACV}
\def\confYear{2026}

\newcommand{\method}{Discrete Facial Encoding\xspace}

\title{\method: A Framework for Data-driven Facial
Display Discovery}

\author{Minh Tran, Maksim Siniukov, Zhangyu Jin, Mohammad Soleymani\\
University of Southern California, Institute for Creative Technologies\\
Los Angeles, CA\\
{\tt\small minhntra@usc.edu}
}

\begin{document}
\maketitle
\begin{abstract}
Facial expression analysis is central to understanding human behavior, yet existing coding systems such as the Facial Action Coding System (FACS) are constrained by limited coverage and costly manual annotation. In this work, we introduce \method (DFE), an unsupervised, data-driven alternative of compact and interpretable dictionary of facial expressions from 3D mesh sequences learned through a Residual Vector Quantized Variational Autoencoder (RVQ-VAE).
Our approach first extracts identity-invariant expression features from images using a 3D Morphable Model (3DMM), effectively disentangling factors such as head pose and facial geometry. We then encode these features using an RVQ-VAE, producing a sequence of discrete tokens from a shared codebook, where each token captures a specific, reusable facial deformation pattern that contributes to the overall expression.
Through extensive experiments, we demonstrate that \method captures more precise facial behaviors than FACS and other facial encoding alternatives. We evaluate the utility of our representation across three high-level psychological tasks: stress detection, personality prediction, and depression detection. Using a simple Bag-of-Words model built on top of the learned tokens, our system consistently outperforms both FACS-based pipelines and strong image and video representation learning models such as Masked Autoencoders. Further analysis reveals that our representation covers a wider variety of facial displays, highlighting its potential as a scalable and effective alternative to FACS for psychological and affective computing applications.
\end{abstract}
    
\section{Introduction}
\label{sec:intro}
Quantitative representation of facial expressions, or facial encoding, is fundamental to psychological and affective computing \cite{zeng2007survey, sariyanidi2014automatic, martinez2012model}. By providing structured and interpretable representations of facial expressions, facial expression coding enables objective analysis of human emotion \cite{pham2019facial}, cognition \cite{craig2008emote}, and behavior \cite{avola2019automatic, hoque2023beamer, kontogiorgos2021systematic}. These representations facilitate the scientific study of social and mental states and enhance the transparency and interpretability of AI applications ranging from clinical diagnostics to human-computer interaction and behavioral health monitoring.

Among facial coding methods, the Facial Action Coding System (FACS) \cite{ekman1978facial} remains the most widely adopted and influential framework. FACS decomposes facial behavior into a standardized set of Action Units (AUs), each corresponding to the activation of a specific facial muscle or group of muscles, thereby enabling principled analysis of how facial patterns relate to underlying psychological processes. Its structured representation has supported a wide range of applications requiring objective interpretation of expressive behaviors \cite{pham2019facial, craig2008emote, avola2019automatic, hoque2023beamer, kontogiorgos2021systematic}. However, traditional FACS coding relies on time-intensive and costly manual annotation, motivating the development of automated AU detection systems \cite{shao2021jaa, song2021uncertain, ijcai2022p173}. Despite recent progress in computer vision, such systems remain limited by moderate accuracy (with state-of-the-art F1 scores typically around 70\% \cite{liu2024norface}) and sensitivity to in-the-wild conditions \cite{yin2024fg}.

To overcome these limitations, we propose a novel data-driven facial expression coding approach utilizing Residual Vector-Quantized Variational Autoencoders \cite{van2017neural, razavi2019generating} (RVQ-VAE). Our method automatically discovers a comprehensive set of expressive facial templates directly from large-scale facial image data \cite{mollahosseini2017affectnet}, enabling complete encoding of observable facial expressions beyond the scope of predefined AU combinations. Unlike FACS-based systems that rely heavily on supervised annotation, our approach is entirely unsupervised, significantly reducing the need for manual labeling and enhancing scalability across diverse datasets. To ensure interpretability and isolate expression-related variations, we operate on 3D Morphable Model (3DMM) features \cite{li2017learning, danvevcek2022emoca}, which allow us to reduce confounding factors such as facial identity (shape) and head pose. We further compress these 3DMM expression features into discrete facial tokens using vector quantization. The resulting discrete facial tokens function as hidden states that influence the reconstructed face. Importantly, each token can be visualized by comparing its associated reconstruction to a neutral template, revealing the specific facial regions it modulates. 

We validate our approach through comprehensive experiments across three key psychological tasks: stress detection \cite{chaptoukaev2023stressid}, personality trait prediction \cite{biel2012youtube}, and depression assessment \cite{ringeval2019avec}. Using a simple Bag-of-Words model \cite{zhang2010understanding} over our learned facial tokens, we demonstrate that our representation consistently outperforms traditional FACS-based features, alternative data-driven facial template discovery systems, and powerful deep image representation learning models such as Masked Autoencoders \cite{ma2024facial, cai2023marlin}. Our analysis shows that the discovered codebook captures a broader and more precise spectrum of facial displays, effectively representing both subtle and complex expressions that are often overlooked by predefined AU-based methods. These findings suggest that learning data-driven facial representations offers a promising and scalable alternative to FACS, opening new avenues for robust, interpretable, and task-relevant facial analysis in psychological and affective computing. Source code and model weights will be released upon publication.

\section{Related work}
\label{sec:related}
\subsection{Facial Action Coding Systems}
Facial action coding is an established visual behavior analysis tool, with the Facial Action Coding System (FACS) \cite{ekman1978facial} serving as the longstanding gold standard. FACS decomposes facial expressions into discrete Action Units (AUs), enabling a structured investigation of the relationship between facial muscle movements and internal states such as emotion, cognition, and mental health.
In psychological and affective computing research, FACS remains widely adopted due to its interpretability and comprehensive coverage of facial expressions \cite{zeng2007survey, sariyanidi2014automatic, martinez2012model}. It allows researchers to quantify facial behavior in a principled and objective manner, facilitating the study of correlations between facial actions and psychological phenomena. 
However, AU Coding traditionally relies on labor-intensive manual coding by certified experts, limiting its scalability. This challenge  has led to increasing interest in automating AU detection \cite{lien1998automated, essa1997coding, bartlett1999measuring, shao2021jaa, song2021uncertain, ijcai2022p173}. Despite recent progress, even state-of-the-art AU detection systems remain imperfect, typically reporting average F1 scores around $0.7$ \cite{liu2024norface}. The task is further complicated by the imbalanced distribution of AUs \cite{kollias2019deep}, where less frequent units are significantly harder to detect accurately \cite{zhang2022transformer}. Most vision-based AU coding systems detect a subset of 44 AUs \cite{chang2024libreface}.
Moreover, current AU detection models exhibit limited generalization across domains \cite{yin2024fg}, often suffering substantial performance drops when evaluated under distribution shifts. These generalization challenges hinder the deployment of AU-based systems in real-world,  critical contexts, suggesting that conventional AU representations may not be sufficiently robust for broad behavioral applications.
\subsection{Interpretable Data-driven Facial Coding}
Given the limitations of Action Units (AU), researchers have explored unsupervised, data-driven representations to capture facial displays more comprehensively. In this direction, Sariyanidi \textit{et al.} propose \textit{Facial Bases} \cite{sariyanidi2017learning}, a method that models facial expressions as linear combinations of localized basis functions, each corresponding to a distinct facial movement (\textit{e.g.}, eyebrow raise). These bases are learned from Gabor phase shifts \cite{fleet1990computation} extracted from facial video sequences, effectively capturing fine-grained temporal motion patterns. The resulting basis coefficients directly reflect movement intensity, enabling the model to represent the gradual evolution of facial expressions over time.
However, since this approach operates on 2D pixel intensities, it struggles to disentangle expression-specific deformations from confounding factors such as head pose, illumination, and facial morphology. As a result, the learned bases may inadvertently encode non-expression-related variations, reducing both interpretability and generalizability, particularly in cross-subject or in-the-wild scenarios.
To address these limitations, the authors extend their framework \cite{sariyanidi2025beyond} by leveraging 3D Morphable Model (3DMM) expression features \cite{sariyanidi2023inequality}, which inherently separate out identity, pose, and lighting variations. This allows the learning process to focus exclusively on expression-related dynamics. Using dictionary learning \cite{mairal2010online} on these 3DMM-derived representations, the method constructs a set of facial bases and derives sparse activation coefficients for each input. These coefficients are then used as features for downstream behavioral prediction tasks, such as autism diagnosis.
Our method differs from Facial Basis in two key ways: first, we leverage deep learning to model the complex, non-linear structure of the 3DMM expression space; and second, we produce a more interpretable, discrete representation by assigning each input to a small set of discrete codebook entries, rather than representing it as a weighted combination of basis templates.
\section{Method}
\begin{figure}[t]
\centering
\scalebox{1.0}{
\includegraphics[width=\linewidth, , trim={0pt 160pt 250pt 30pt}, clip]{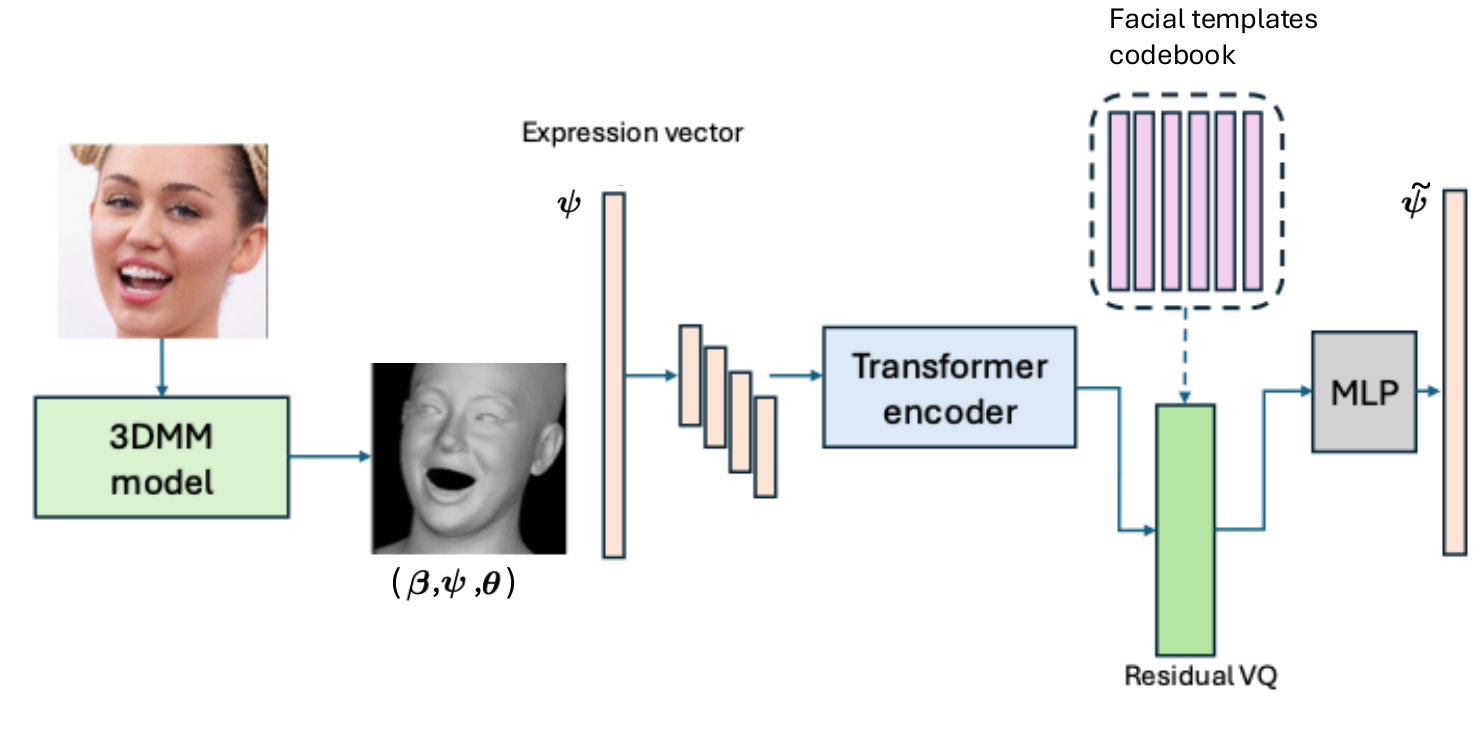}
}
\caption{Overview of our proposed expression coding framework. Given an input expression vector \( \boldsymbol{\psi} \) extracted from a 3DMM model, a transformer-based encoder maps it into a latent representation. This representation is then quantized using Residual VQ to produce discrete expression tokens. A lightweight MLP decoder reconstructs the expression vector \( \tilde{\boldsymbol{\psi}} \), preserving additive structure and interpretability.}
\label{fig:overview_fig}
\vspace{-10pt}
\end{figure}
\textcolor{black}{An overview of our proposed model is available in Figure \ref{fig:overview_fig}. Given a face image, our goal is to decompose the expression into interpretable, discrete components. To this end, we first reduce the influence of identity and other factors such as face shape and head pose by extracting expression parameters using a 3D Morphable Model (3DMM) \cite{danvevcek2022emoca}. 3DMMs are designed to disentangle expression from identity, and their expression parameters mostly contain information about expression (they may contain residual identity information due to their limitations). 
We then encode these 3DMM expression vectors using a Residual Vector-Quantized Variational Autoencoder (RVQ-VAE) \cite{van2017neural, razavi2019generating}, which maps each input to a set of discrete tokens. These tokens provide a compact and interpretable representation of facial behavior, and can be visualized by decoding them back into facial expressions using the 3DMM.}
\subsection{Feature Extraction with EMOCA}
\textcolor{black}{We use EMOCA \cite{danvevcek2022emoca} to extract expression features from facial images due to its optimization to capture expressions of emotions, popularity in facial behavior generation \cite{ng2023can, tran2024dim} and its straightforward process for converting expression vectors into facial meshes.}
EMOCA is a 3D face reconstruction model built on top of the DECA 3D Morphable Model \cite{feng2021learning}, which represents a face as a deformation of a neutral template mesh \( \mathbf{T} \in \mathbb{R}^{3 \times N} \), where \( N \) is the number of vertices. The final mesh \( \mathbf{M} \) is computed as:
\begin{equation}
\mathbf{M} = W\left( \mathbf{T} + B_s(\boldsymbol{\beta}) + B_e(\boldsymbol{\psi}), J(\boldsymbol{\beta}), \boldsymbol{\theta} \right)
\end{equation}
where \( \boldsymbol{\beta} \), \( \boldsymbol{\psi}  \), and \( \boldsymbol{\theta}\) are the shape, expression, and pose vectors, respectively. \( B_s \) and \( B_e \) are the shape and expression blendshape functions, \( J(\boldsymbol{\beta}) \) defines how to compute joint locations from mesh vertices, $\mathbf{T}$ is the ``zero pose" template mesh, and \( W(\cdot, J, \boldsymbol{\theta}) \) is the linear blend skinning function that applies pose-dependent deformations. 

EMOCA improves upon DECA by enhancing the expressivity of reconstructed faces. It introduces an emotion consistency loss, which encourages the emotion features of the input image to match those of the rendered reconstruction. Specifically, the model minimizes the mean squared error (MSE) between emotion embeddings extracted from the input and the rendered image. This regularization helps EMOCA better preserve the emotional content and subtle expressive details of the original input.
Given the high emotional fidelity of EMOCA and its strong ability to disentangle expression from identity and pose, we use the extracted expression parameters as the input to our framework.

\subsection{Discrete Expression Encoding with Residual VQ-VAE}
Our model maps the EMOCA expression vector into a discrete latent code using a Residual Vector-Quantized Variational Autoencoder (RVQ-VAE) \cite{razavi2019generating}, combining transformer-based encoding and multi-stage residual quantization.\\
\textbf{Encoder.} We reshape the input expression vector into a sequence of \(T\) tokens of dimension \(d\). Each token is projected into a hidden space via a linear layer, then passed through a Transformer encoder \cite{vaswani2017attention}. The output is mean-pooled and projected into a latent vector \( \mathbf{z}_0 \in \mathbb{R}^{D} \).\\
\textbf{Residual Quantization.} We apply residual quantization over \( L \) stages using a shared codebook  \( Q \in \mathbb{R}^{K \times D} \) with \( K \) entries. At each stage \( i \), we quantize the residual:
\begin{align}
k_i &= \arg\min_{k} \| \mathbf{z}_{i-1} - \mathbf{e}_k \|_2^2 \\
\mathbf{z}_i &= \mathbf{z}_{i-1} - \mathbf{e}_{k_i}
\end{align}
This process converts \( \mathbf{z}_0 \) into a sequence of $L$ discrete tokens with an additive property. The first token represents the face template most similar to the given facial input, and each subsequent token encodes finer residual details, progressively refining the facial representation. The final quantized vector is the sum of selected codes: $\mathbf{z_q} = \sum_{i=1}^{L} \mathbf{e}_{k_i}$.\\
\textbf{Decoder.} Unlike traditional VQ-VAEs \cite{van2017neural} that have symmetric encoder and decoder architectures, our decoder consists of a simple linear projection layer that maps the quantized latent code $z_q$ back to the input dimension. 
\begin{equation}
\hat{\boldsymbol{\psi}} = g_{\theta}(\hat{\mathbf{z}})
\end{equation}
\begin{figure}[t]
\centering
\scalebox{0.6}{
\includegraphics[width=\linewidth, trim={160pt 210pt 460pt 60pt}, clip]{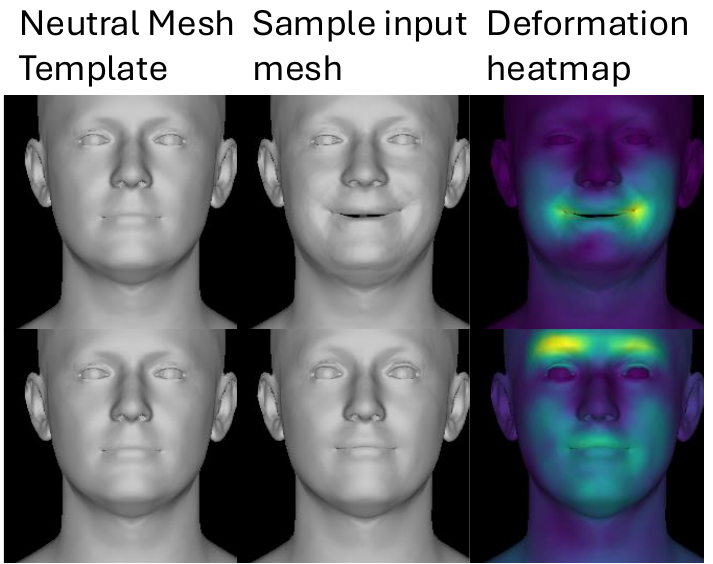}
}
\caption{Some examples of deformation heatmap.}
\label{fig:interpet_pipeline}
\vspace{-15pt}
\end{figure}
This design is motivated by two factors: 1) the additive structure of the architecture enhances interpretability when visualizing the contributing components of a facial display, and 2) the encoder's representation is the primary focus during training. A more complex decoder does not significantly improve the encoder's ability to learn a rich latent representation, as the decoder's role is primarily to reconstruct the input once the latent space has been learned. \\
\textbf{Training losses.} Our model's training objective includes a reconstruction loss and a commitment loss, following the formulation introduced in the original VQ-VAE framework \cite{vaswani2017attention}. The reconstruction loss ensures fidelity between the input and the output, while the commitment loss encourages the encoder outputs to remain close to their assigned codebook vectors:
\begin{equation}
\mathcal{L}_{vq} = 
{\| \boldsymbol{\psi} - \hat{\boldsymbol{\psi}} \|_2^2}
+ \lambda_{\text{commit}} \sum_{i=1}^{L} \| \mathbf{z}_i - \text{sg}(\mathbf{e}_{k_i}) \|_2^2
\end{equation}
To encourage the model to capture fine-grained and localized facial details, we incorporate two regularization terms during training: an \( \ell_1 \)-penalty and an orthogonality loss. The \( \ell_1 \)-penalty promotes sparsity in the codebook usage, encouraging each token to specialize in distinct facial regions. Meanwhile, the orthogonality loss ensures that the decoded codebook embeddings remain diverse and non-redundant. It is defined as:
\begin{equation}
\mathcal{L}_{\text{orth}} = \frac{1}{K(K - 1)} \sum_{i \neq j} \left( \mathbf{e}_i^\top \mathbf{e}_j \right)^2
\end{equation}
where \( \mathbf{e}_i \in \mathbb{R}^d \) is the embedding of the \(i\)-th codebook entry, and \( K \) is the size of the codebook. Overall, our model training objective is
\begin{equation}
\mathcal{L} = 
\mathcal{L}_{vq} + \lambda_{\text{orth}} \mathcal{L}_{\text{orth}} + \lambda_{\text{reg}} {\| \hat{\boldsymbol{\psi}} \|_1}
\end{equation}

\subsection{Visualization of facial templates}
After training, each input expression is represented as a discrete code sequence \( [k_1, k_2, \dots, k_L] \), providing a compact and interpretable tokenization of facial expressions. Each token corresponds to a quantized latent vector that can be decoded into a 3D facial mesh using the 3DMM decoder \cite{danvevcek2022emoca}, enabling direct visual inspection. Since the 3DMM effectively disentangles expression from identity, pose, and lighting, we can manipulate only the expression coefficients while keeping other factors fixed. This allows us to isolate and visualize the specific facial deformation induced by each token in a controlled manner.
\begin{figure}[t]
\centering
\scalebox{1.0}{
\includegraphics[width=0.9\linewidth, trim={100pt 250pt 280pt 60pt}, clip]{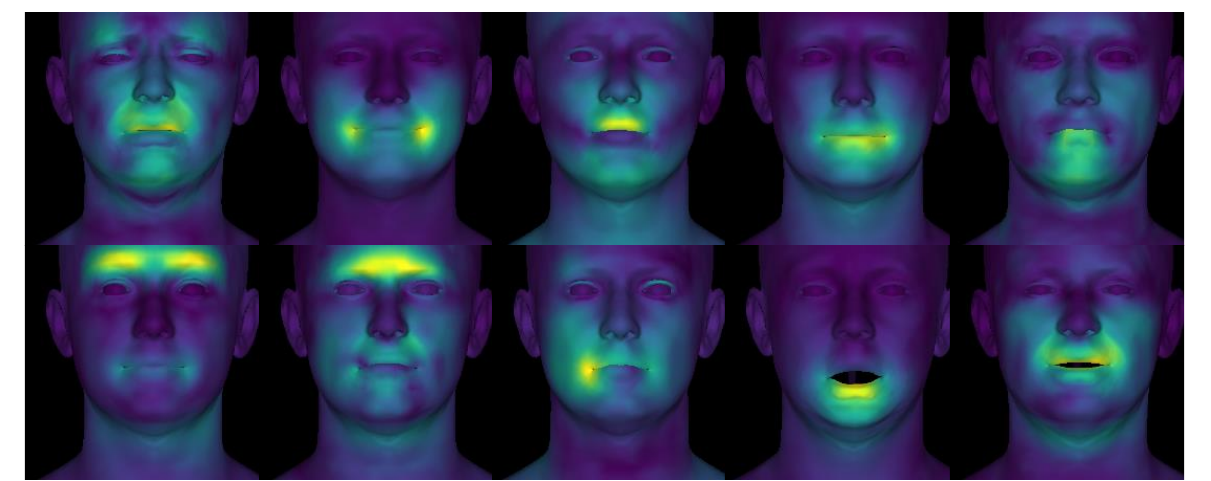}
}
\caption{Some expression templates discovered by our system.
}
\label{fig:sample_templates}
\vspace{-10pt}
\end{figure}
\begin{figure}[t]
\centering
\scalebox{1.0}{
\includegraphics[width=0.9\linewidth, trim={180pt 20pt 180pt 20pt}, clip]{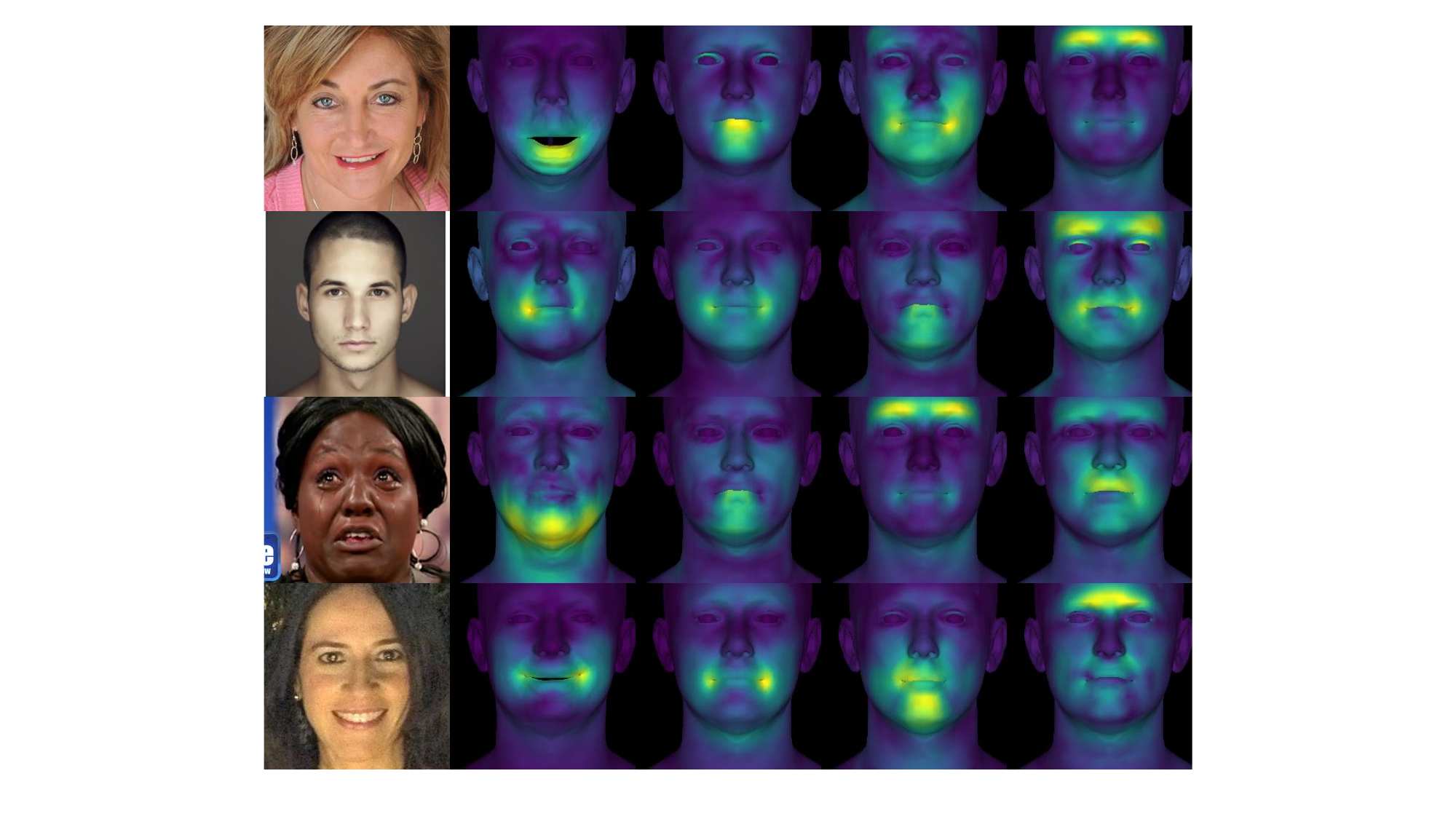}
}
\caption{Example facial images and their corresponding token decompositions produced by our system.}
\label{fig:decompose_sample}
\vspace{-15pt}
\end{figure}

Specifically, to visualize how each discrete token contributes to facial geometry, we render a \textit{deformation heatmap} by comparing the reconstructed 3D mesh against a neutral face template (\textit{i.e.}, \( \boldsymbol{\psi} = \mathbf{0} \)). For each discrete code, we decode it via EMOCA to obtain the reconstructed face mesh. We then compute a per-vertex Euclidean distance between the reconstructed mesh and the neutral mesh:
\[
\mathbf{d}_v = \left\| \mathbf{v}_v - \mathbf{v}_v^{\text{ref}} \right\|_2, \quad \text{for } v = 1, \dots, N,
\]
where \( \mathbf{v}_v \in \mathbb{R}^3 \) is the position of vertex \( v \) in the reconstructed mesh, and \( \mathbf{v}_v^{\text{ref}} \) is the corresponding vertex in the neutral mesh. These distances are normalized and mapped to a perceptual colormap to produce interpretable heatmaps that highlight localized expression-driven deformations.

We demonstrate examples of our interpretability pipeline in Figure~\ref{fig:interpet_pipeline}. As shown, it is often difficult to determine which facial regions are activated by simply inspecting the reconstructed face mesh. However, by comparing it with the neutral face template and visualizing the deformation as a heatmap, we can clearly localize the regions influenced by each token, offering a more interpretable and spatially grounded understanding of the token's effect on facial expression. In Figure~\ref{fig:sample_templates}, we show several example expressions encoded by our system. The results indicate that the model effectively captures diverse expression patterns, with different tokens corresponding to localized facial deformations that resemble distinct combinations of muscle activations. Finally, Figure~\ref{fig:decompose_sample} illustrates how the expression of a given input image can be decomposed into a set of additive components, demonstrating both the model's accuracy and the interpretability of its token-based representation.

\section{Experiments}
We compare the proposed facial expression coding system with existing approaches along three key dimensions: (1) accuracy in preserving facial expressions, (2) utility as a feature representation for downstream psychological tasks, and (3) diversity in capturing a wide range of facial expressions.

\subsection{Datasets}
\vspace{-10pt}
\textcolor{black}{
\subsubsection{Pre-training Dataset} We pre-train our RVQ-VAE model on the AffectNet \cite{mollahosseini2017affectnet}, a large-scale collection of approximately 350K face images annotated with both categorical and dimensional emotion labels. AffectNet offers substantial variability in appearance, expression, ethnicity and pose, making it well-suited for learning robust and generalizable expression representations. 
\subsubsection{Evaluation Datasets} For evaluation, we use several datasets, organized by their specific purposes:\\
\noindent \textbf{Expression Preservation and Diversity:}
\begin{itemize}
    \item Aff-Wild2 \cite{kollias2019deep}: This in-the-wild video dataset is annotated with frame-level Action Unit (AU) labels, providing a rich and dynamic set of facial behaviors. Its scale and expressive variety enable a comprehensive assessment of how well different coding systems capture subtle and complex facial motions.
    \item SmileStimuli \cite{martin2021evidence}: This dataset contains 45 video recordings from 15 professional actors, each portraying one of three categories of smiles (dominance, affiliation or reward). The balanced distribution of smile types allows us to further explore the diversity of expressions captured by our system.
\end{itemize}
\noindent \textbf{Downstream Psychological Tasks:}
\begin{itemize}
    \item Stress Identification \cite{chaptoukaev2023stressid}: We evaluate on the StressID dataset \cite{chaptoukaev2023stressid}, a recent multimodal benchmark designed to assess stress levels in real-world human interactions. The dataset comprises over 1,200 annotated video segments collected from 65 participants undergoing stress-inducing conditions such as cognitive load and public speaking. Each segment is rated on a perceived stress scale from 1 to 10 and subsequently converted into binary or three-class stress labels. Given the presence of rich facial expressions throughout the recordings, this dataset is well-suited for testing facial encoding systems.
    \item Depression Detection \cite{ringeval2019avec}: We use the dataset from the AVEC 2019 challenge \cite{ringeval2019avec}, which targets automatic depression analysis. Specifically, we consider two tasks: (1) depression severity regression, and (2) binary classification of depressed vs. non-depressed subjects. The dataset consists of video recordings from 275 subjects, totaling approximately 73 hours of audiovisual data. Each subject underwent a semi-structured clinical interview conducted by a virtual agent, with depression severity assessed using the PHQ-8 questionnaire. The interviews were performed in a Wizard-of-Oz (WoZ) setup, where the virtual agent was controlled by a human operator.
    \item ChaLearn First Impressions \cite{biel2012youtube}: We use the ChaLearn First Impressions dataset \cite{biel2012youtube}, a large-scale benchmark for apparent personality recognition from short videos. It contains over 10,000 video segments featuring individuals speaking in unconstrained settings, each annotated with continuous scores for the Big Five personality traits: openness, conscientiousness, extraversion, agreeableness and neuroticism. These scores range from 0 to 1, indicating the perceived strength of each trait.
\end{itemize}
}
\subsection{Baselines}
Our primary baseline is the widely used \textbf{Facial Action Unit} (AU) system \cite{ekman1978facial}. For datasets lacking annotated AU labels, we use LibreFace \cite{chang2024libreface} to extract AU features. \textcolor{black}{We compare our method with automatically tracked (rather than human-coded) AU features, as both approaches are automated and do not require human input in the pipeline.}
In addition, we include \textbf{Facial Basis} \cite{sariyanidi2025beyond} as a baseline for evaluating utility in downstream psychological tasks. However, due to its continuous, non-discrete representation, we do not include it in experiments focused on facial accuracy preservation or expression diversity.

To contextualize our method's performance in broader representation learning, we also report results from popular image and video encoding models, including \textbf{MAE-Face} \cite{ma2024facial}, \textbf{VideoMAE} \cite{tong2022videomae}, and \textbf{MARLIN} \cite{cai2023marlin}. While these models are not designed for interpretability, they serve as strong representation learning baselines for assessing utility in psychological inference tasks. Their inclusion highlights the trade-off between interpretability and raw representational power in modern deep learning approaches.
\subsection{Implementation details}
Our model is implemented in PyTorch and trained on a single NVIDIA H100 GPU. 
The input to the model is a 3DMM expression vector reshaped to size $T=10$ and $d=5$.
Our transformer encoder contains $6$ layers with $4$ attention heads, and a hidden dimension $\mathcal{D}=128$. We apply residual vector quantization (RVQ) with \( L = 4 \) quantization stages and a shared codebook \( \mathcal{C} \in \mathbb{R}^{K \times D} \) of size \( K = 64 \) and \( D = 50 \). 
The decoder \( f_{\text{dec}}: \mathbb{R}^{D} \rightarrow \mathbb{R}^{50} \) is a single linear projection that reconstructs the original expression vector. 
We set hyperparameters as follows: \( \beta = 0.25 \), \( \lambda_{\text{orth}} = 1.0 \), \( \lambda_{\text{sparse}} = 0.1 \), and \( \lambda_1 = 1 \times 10^{-4} \). \textcolor{black}{We provide ablation studies on our hyper-parameter choices in the supplemental materials.}
The model is trained using the Adam optimizer \cite{kingma2014adam} with learning rate \( 1 \times 10^{-4} \), batch size $512$ for $500$ epochs. 
\textcolor{black}{For downstream psychological tasks with videos, we represent each video using a Bag-of-Words (BoW) approach, encoding it as a frequency distribution over the codebook entries. Detailed modeling procedures for each downstream task are provided in the corresponding discussion sections.}

\section{Discussion}

\begin{figure}[b]
\centering
\scalebox{1.0}{
\includegraphics[width=\linewidth, trim={50pt 20pt 350pt 0pt}, clip]{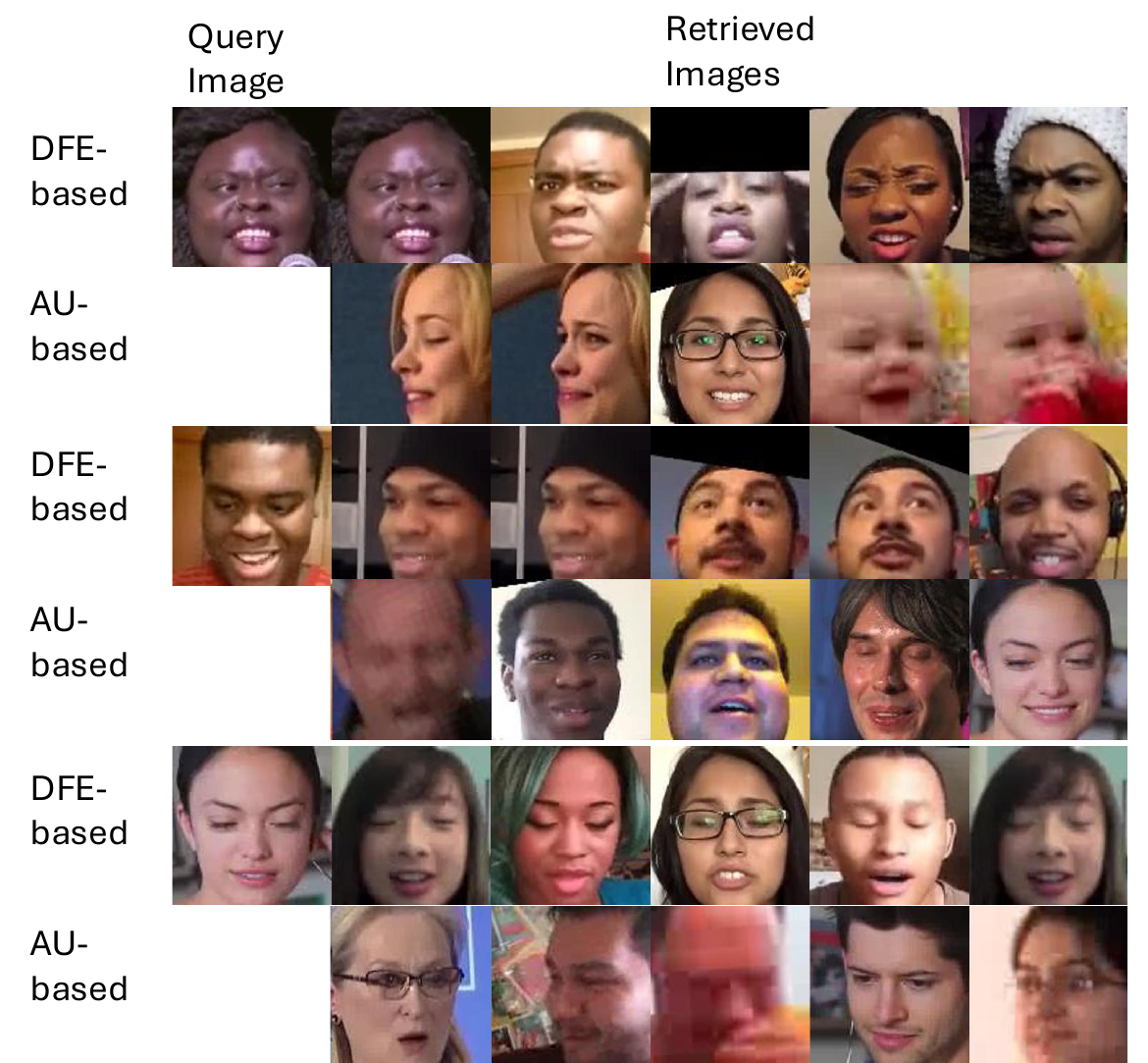}
}
\caption{Qualitative retrieval examples comparing our token-based representation (DFE) with AU-based encoding.}
\label{fig:retrieval_samples}
\vspace{-20pt}
\end{figure}
\subsection{Evaluating Expression Accuracy}
To assess how accurately our VQ-VAE-based encoding system captures facial expressions, we conduct a retrieval-based evaluation against a baseline system based on Facial Action Units (AUs). The goal of this experiment is to evaluate how well the learned representations preserve expression-relevant information.

Given a query facial image, we use both the human-annotated AU-based and VQ-VAE-based systems to retrieve visually similar samples from a large database consisting of 300K frames from a subset of the Aff-Wild2 dataset \cite{kollias2019deep} with human-annotated AU labels. Due to the large scale of Aff-Wild2 \cite{kollias2019deep}, we randomly subsampled overlapping short clips of 64 frames to construct this retrieval set. For both systems, each image is first converted into a binary encoding vector: for the AU-based system, each element indicates whether a specific AU is activated; for the VQ-VAE system, each element indicates whether a discrete token is present in the coded sequence. Using these binary vectors, we retrieve all database images that have the exact same encoding as the query. If fewer than five matches are found, the query is excluded from evaluation.

To fairly assess the quality of retrievals, we employ SMIRK \cite{retsinas20243d}, a recently introduced 3DMM-based system for extracting expression features from both the query and retrieved images. We intentionally avoid using EMOCA \cite{danvevcek2022emoca} to prevent evaluation bias, as our model is trained to reconstruct EMOCA-derived features. Instead, SMIRK-derived expression vectors serve as a neutral ground truth for measuring retrieval quality, allowing us to focus exclusively on expression similarity while disregarding confounding factors such as head pose and identity. Additionally, we use MAE-Face \cite{ma2024facial}, a state-of-the-art self-supervised facial representation model, to extract features from the same retrieval sets. Unlike 3DMM-based encodings, MAE-Face \cite{ma2024facial} captures holistic facial representations that may include irrelevant attributes such as head pose and identity, providing useful additions to our expression-focused evaluation.

We report three quantitative metrics: \textbf{(1) Mean Euclidean Distance / Cosine Similarity}: Measures the average distance between the SMIRK expression vector of the query and those of the retrieved images. Lower values for Euclidean distance and higher values for Cosine Similarity indicate more accurate expression matching. \textbf{(2) Standard Deviation}: Computes the average standard deviation of the SMIRK expression vectors within each retrieval group. Lower values suggest that the system captures more fine-grained and consistent expression features.
\begin{table}[t]
\centering
\caption{Retrieval accuracy comparison between AU-based and VQ-VAE-based encodings evaluated using SMIRK \cite{retsinas20243d} and MAE-Face \cite{ma2024facial} features. CosSim = average cosine similarity; EucDist = average Euclidean distance; Std = average standard deviation.}
\resizebox{\linewidth}{!}{
\begin{tabular}{llccc}
\toprule
\textbf{Evaluation} & \textbf{Method} & \textbf{CosSim} $\uparrow$ & \textbf{EucDist} $\downarrow$ & \textbf{Std} $\downarrow$ \\
\midrule
\multirow{2}{*}{SMIRK} 
    & AU-based        & 0.5491 & 8.7318 & 0.7263 \\
    & Ours (VQ-VAE)   & \textbf{0.8184} & \textbf{4.9599} & \textbf{0.4436} \\
\midrule
\multirow{2}{*}{MAE-Face} 
    & AU-based        & 0.9821   & 5.0354   & 0.0950   \\
    & Ours (VQ-VAE)   & \textbf{0.9913} & \textbf{3.0115} & \textbf{0.0626} \\
\bottomrule
\end{tabular}
}
\label{tab:retrieval_accuracy}
\end{table}

The quantitative results in Table~\ref{tab:retrieval_accuracy} and qualitative results in Figure~\ref{fig:retrieval_samples} demonstrate that our VQ-VAE-based representation significantly outperforms the AU-based encoding in all retrieval metrics. The higher cosine similarity and lower Euclidean distance indicate that our method retrieves samples with more accurate expression matches. Additionally, the reduced standard deviation shows that our system captures more consistent and fine-grained expression variations across retrieved sets.

\subsection{Evaluating Expression Diversity}
To quantify the diversity of expressions captured by each coding system, we compute the normalized entropy over their respective feature vocabularies. We conduct this analysis on a fixed subset of approximately 300K video frames from the Aff-Wild2 dataset \cite{kollias2019deep} with human-annotated AU labels. For each system, we count the frequency of occurrence for each discrete unit—AUs in the baseline system and tokens in our VQ-VAE model.

Let \( \mathbf{p} = [p_1, p_2, \dots, p_K] \) denote the empirical distribution over a vocabulary of size \( K \), we first compute the entropy of \( \mathbf{p} \).
To account for different vocabulary sizes across systems, we normalize the entropy by dividing by the dimensionality \( K \), yielding the \emph{normalized entropy}:

\begin{equation}
\hat{H}(\mathbf{p}) = \frac{H(\mathbf{p})}{K} = \frac{= -\sum_{i=1}^{K} p_i \log_2 p_i}{K}
\end{equation}
A low normalized entropy indicates that the system predominantly activates a small subset of features (collapse), leading to poor expressive coverage and redundancy. In contrast, higher entropy implies that the system utilizes a broader range of features across inputs, suggesting greater expressiveness and diversity in facial representation.
\begin{table}[t]
\centering
\caption{Quantitative diversity comparison of facial encoding systems. Higher entropy indicates broader usage of distinct tokens, while lower NMI suggests less redundant features.}
\resizebox{0.7\linewidth}{!}{
\begin{tabular}{lcc}
\toprule
\textbf{Method} & \textbf{Entropy} $\uparrow$ & \textbf{NMI} $\downarrow$ \\
\midrule
AU-based (human)       & 0.846 & 0.061  \\
AU-based (auto)       & 0.913 & 0.080  \\
Ours (VQ-VAE)   & \textbf{0.926} & \textbf{0.004}  \\
\bottomrule
\end{tabular}
}
\vspace{-10pt}
\label{tab:diversity_quant}
\end{table}



As a second measure of expression diversity, we assess the redundancy among features in each encoding system by computing the \textit{average normalized mutual information (NMI)} between features. Mutual information \cite{shannon1948mathematical} quantifies the amount of shared information between two variables, while the normalization accounts for differences in feature entropy, making the measure directly comparable across different encoding systems \cite{strehl2002cluster}. The key intuition is that lower normalized mutual information suggests more independent and disentangled features, indicating a more expressive representation \cite{peng2005feature}.

Given an encoding matrix \( \mathbf{X} \in \mathbb{R}^{N \times K} \), where each row is a binary vector of feature activations, we compute the normalized mutual information between all unique feature pairs (columns of \( \mathbf{X} \)). The average normalized mutual information is then calculated as:
\begin{equation}
\text{avg NMI} = \frac{2}{K(K - 1)} \sum_{i < j} \frac{I(X_i; X_j)}{\sqrt{H(X_i) H(X_j)}}
\end{equation}
where \( I(X_i; X_j) \) is the mutual information between feature \( i \) and feature \( j \), and \( H(X_i) \) is the entropy of feature \( i \). A lower value of \( \text{avg NMI} \) indicates that features tend to vary independently across samples, reflecting higher diversity and lower redundancy. Conversely, a higher NMI suggests that many features are co-activated and share overlapping information.

We provide the quantitative results of the two diversity metrics in Table~\ref{tab:diversity_quant}. Our VQ-VAE representation achieves the highest normalized entropy (0.926), indicating that it activates a broader range of tokens across samples compared to both manually and automatically extracted AUs. Furthermore, it exhibits the lowest average normalized mutual information (0.004), suggesting that the learned tokens are highly independent and minimally redundant. Together, these results demonstrate that our system achieves superior diversity with minimal redundancy, offering a more expressive and disentangled facial representation.
\begin{table}[t]
\centering
\caption{Smile-type classification performance on the SmileStimuli dataset \cite{martin2021evidence}. All values are multiplied by $100$ for readability.}
\resizebox{0.5\linewidth}{!}{
\begin{tabular}{lccc}
\toprule
\textbf{Method} & \textbf{AUC} & \textbf{Acc} & \textbf{F1} \\
\midrule
AU-based                & 69.2              & 51.2                   & 50.2                   \\
Ours         & \textbf{71.4}     & \textbf{58.1}          & \textbf{59.4}          \\
\bottomrule
\end{tabular}
\label{tab:smile}

}
\vspace{-10pt}
\end{table}

\begin{figure}[t]
\centering
\scalebox{1.0}{
\includegraphics[width=0.9\linewidth, trim={20pt 50pt 350pt 0pt}, clip]{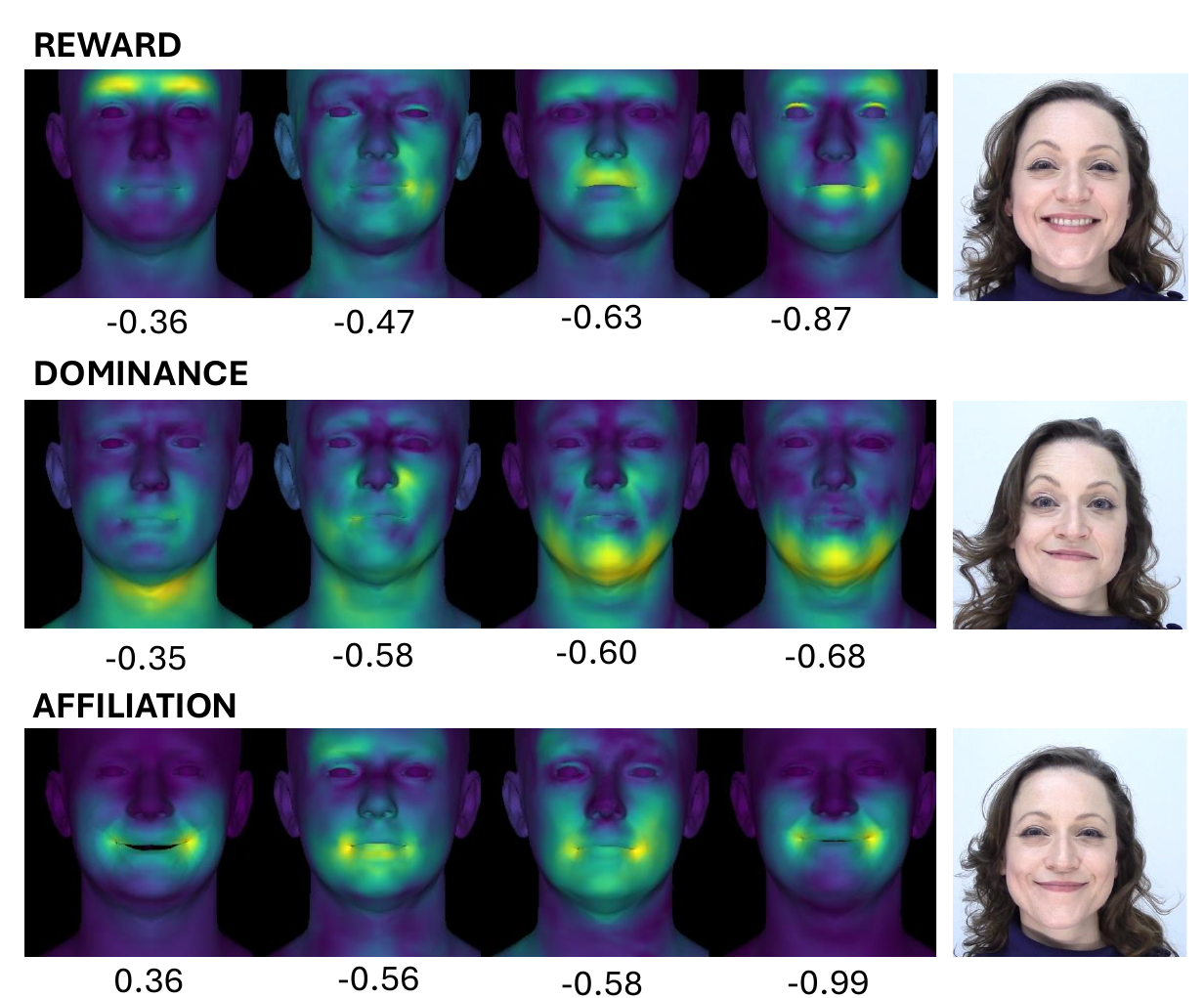}
}
\caption{Top-4 discriminative templates for smile classification.}
\label{fig:smile_fig}
\vspace{-5pt}
\end{figure}

Finally, we validate the diversity of expressions captured by our system using the SmileStimuli dataset \cite{martin2021evidence}, which contains posed smiles categorized into dominance, affiliation, and reward smiles. Given the limited size of the dataset ($45$ samples in total), we employ a Logistic Regression model with a leave-one-out cross-validation strategy. We compare the performance of models using our VQ-based token representations against models using traditional Action Unit (AU) features. The results, summarized in Table~\ref{tab:smile}, show that our system consistently outperforms the AU-based model across all evaluation metrics, including Accuracy, F1 Score, and AUC. This demonstrates that our learned token representations offer stronger discriminative power for differentiating subtle social smiles, involving asymmetry. However, it is important to note that our system encodes only geometric information, and prior research suggests that geometry alone may not fully capture facial expressions \cite{thorstenson2018emotion}, which may explain the imperfect performance. To further illustrate the interpretability of our system, we visualize the most important facial templates—identified based on the log of the absolute values of the learned logistic regression coefficients—for each smile type. The top-4 templates are shown in Figure~\ref{fig:smile_fig}, highlighting the diversity of expressions captured by our approach.

\subsection{Evaluating Feature Utility}
We evaluate the usefulness of the learned tokens on three downstream high-level psychological tasks: depression detection, stress identification, and personality trait prediction.
\begin{table*}[t]
\centering
\caption{Performance across five personality dimensions using Accuracy and CCC scores. All values are multiplied by $100$ for readability.}
\resizebox{0.8\linewidth}{!}{
\begin{tabular}{lcccccccccc}
\toprule
\multirow{2}{*}{\textbf{Model}} & \multicolumn{2}{c}{\textbf{Openness}} & \multicolumn{2}{c}{\textbf{Conscientiousness}} & \multicolumn{2}{c}{\textbf{Extraversion}} & \multicolumn{2}{c}{\textbf{Agreeableness}} & \multicolumn{2}{c}{\textbf{Neuroticism}} \\
\cmidrule(r){2-3} \cmidrule(r){4-5} \cmidrule(r){6-7} \cmidrule(r){8-9} \cmidrule(r){10-11}
& Acc $\uparrow$ & CCC $\uparrow$ & Acc $\uparrow$ & CCC $\uparrow$ & Acc $\uparrow$ & CCC $\uparrow$ & Acc $\uparrow$ & CCC $\uparrow$ & Acc $\uparrow$ & CCC $\uparrow$ \\
\midrule
FaceMAE            & 88.3 & 21.7 & 87.8 & 29.6 & 88.1 & 35.6 & 89.6 & 9.5 & 87.8 & 19.1 \\
Marlin             & 88.8 & 18.8 & 87.7 & 35.5 & 87.9 & 22.2 & 88.6 & 21.0 & 88.0 & 20.5 \\
VideoMAE       & 88.9 & 27.2 & 87.7 & 35.7 & 88.3 & 23.3 & 89.4 & 22.6 & 88.1 & 25.9 \\ \hline
AU & 89.8 & 38.0 & 88.8 & 35.1 & 90.0 & 45.7 & 90.3 & 25.8 & 89.2 & 36.3 \\
Facial Basis   & 89.9 & 37.2 & 88.7 & 31.4 & 90.0 & 46.7 & 90.6 & 22.7 & 89.1 & 33.1 \\
Ours (VQ-VAE)  & \textbf{90.2} & \textbf{43.1} & \textbf{89.2} & \textbf{40.0} & \textbf{90.5} & \textbf{53.8} & \textbf{90.8} & \textbf{30.5 }& \textbf{89.6} & \textbf{42.0 }\\
\bottomrule
\end{tabular}
\label{tab:personality}
}
\vspace{-10pt}
\end{table*}
\begin{table}[t]
\centering
\caption{Performance comparison on the AVEC 2019 Depression Detection task. We report RMSE and CCC for the regression subtask, and Accuracy and AUC for the binary classification subtask. All values are multiplied by $100$ for readability.}
\resizebox{0.8\linewidth}{!}{
\begin{tabular}{lcccc}
\toprule
\textbf{Model} & \textbf{RMSE} $\downarrow$ & \textbf{CCC} $\uparrow$ & \textbf{Acc} $\uparrow$ & \textbf{AUC} $\uparrow$ \\
\midrule
FaceMAE            & 8.4 & 6.0 & 61.1 & 54.1 \\
Marlin             & 7.6 & \textbf{19.8} & 59.3 & 52.8 \\
VideoMAE       & 7.7 & 10.4 & 61.1 & 56.1 \\ \hline
AU (LibreFace) & 8.3 & 14.1 & \textbf{67.9} & 62.4 \\
Facial Basis   & 7.5 & 8.2 & \textbf{67.9} & 60.0 \\
Ours (VQ-VAE)  & \textbf{7.2} & \textbf{19.8} & \textbf{67.9} & \textbf{63.3} \\
\bottomrule
\end{tabular}
}
\label{tab:depression}
\vspace{-10pt}
\end{table}

\begin{table}[t]
\centering
\caption{Performance comparison on the StressID dataset. We report F1 Score and Balanced Accuracy for both binary and multi-class classification. All values are multiplied by $100$ for readability.}
\resizebox{0.8\linewidth}{!}{
\begin{tabular}{lcccc}
\toprule
\multirow{2}{*}{\textbf{Model}} & \multicolumn{2}{c}{\textbf{Binary }} & \multicolumn{2}{c}{\textbf{Multiclass}} \\
 & \textbf{F1} $\uparrow$ & \textbf{BAcc} $\uparrow$ & \textbf{F1} $\uparrow$ & \textbf{BAcc} $\uparrow$ \\
\midrule
FaceMAE            & 56.2 & 58.4 & 40.2 & 40.7 \\
Marlin             & 59.6 & 59.5 & 49.8 & 50.3 \\
VideoMAE       & 72.3 & 65.1 & 45.2 & 45.7 \\ \hline
AU (LibreFace) & 70.0 & 70.0 & 55.0 & 55.0 \\
Facial Basis   & 72.2 & 71.9 & 58.5 & 57.8 \\
Ours (VQ-VAE)   & \textbf{73.3} & \textbf{72.9} & \textbf{61.1} & \textbf{60.3}\\
\bottomrule
\end{tabular}
}
\label{tab:stress}
\vspace{-20pt}
\end{table}
\textcolor{black}{For Depression Detection, we follow the official AVEC 2019 evaluation protocol with train/validation/test splits and report two standard metrics: Root Mean Square Error (RMSE) and Concordance Correlation Coefficient \cite{lawrence1989concordance} (CCC) for the regression task. For the binary classification task, we report accuracy and AUC score. For Stress Identification, we follow the official evaluation protocol and report both F1-score and balanced accuracy. For Personality Detection, following prior work, we report two metrics: the Concordance Correlation Coefficient (CCC) and Accuracy, defined as $1-MAE$, where $MAE$ denotes the mean absolute error. We use Support Vector Machines (SVM) for classification tasks and Support Vector Regression (SVR) for regression tasks.}

We present the results for personality detection in Table~\ref{tab:personality}, depression detection in Table~\ref{tab:depression}, and stress identification in Table~\ref{tab:stress}. Across all tasks, our proposed method consistently outperforms baseline approaches, demonstrating the effectiveness and generalizability of our discrete token representation. Notably, our model surpasses even end-to-end, non-interpretable image and video representation learning models, indicating that the learned tokens are not only compact and interpretable but also semantically rich and highly informative for psychological inference.
\section{Limitation}
While our method offers interpretable and effective facial expression encoding, it has several limitations. First, the quality and expressivity of our learned tokens are inherently dependent on the richness of the 3DMM features used during training; limited or biased 3DMM expression representations may constrain the model’s capacity. \textcolor{black}{Furthermore, 3DMM features may still contain residual identity information, which can limit the effectiveness of our method in modeling fully identity-independent facial templates. Addressing this limitation and further reducing identity leakage remains an important direction for future work.} Second, although our framework is currently developed and evaluated on static images, it is naturally extensible to video inputs by incorporating temporal modeling—an avenue we leave for future work. Third, the facial display templates are biased by the dataset on which the RQ-VAE is trained, which might not capture all cultural and individual variations. Fourth, our model ignores skin color changes that contain information about human inner states \cite{thorstenson_emotion-color_2018}. Finally, our current model focuses solely on facial expression and does not account for other behavioral cues such as eye gaze, head pose, or body movement, which are often critical in psychological and affective understanding.
\section{Conclusion}
We introduced a novel framework for interpretable facial expression encoding using a VQ-VAE architecture trained on 3DMM-derived expression features. By representing facial expressions as discrete token sequences, our method enables both semantic interpretability and effective downstream use in psychological applications. 
Through comprehensive experiments, we demonstrate that our learned tokens outperform existing facial encoding systems—including Action Units and recent self-supervised models—across metrics of expression fidelity, feature diversity, and predictive utility. Furthermore, our approach offers a structured and visualizable representation space, bridging the gap between human-interpretable codes and machine-learned representations. Future work will explore integrating temporal dynamics and multimodal signals such as gaze and head movement to enhance behavioral modeling.
{
    \small
    \bibliographystyle{ieeenat_fullname}
    \bibliography{main}
}
\appendix
\appendix


\begin{figure*}[ht]
\centering
\scalebox{1.0}{
\includegraphics[width=0.9\linewidth, trim={0pt 0pt 330pt 0pt}, clip]{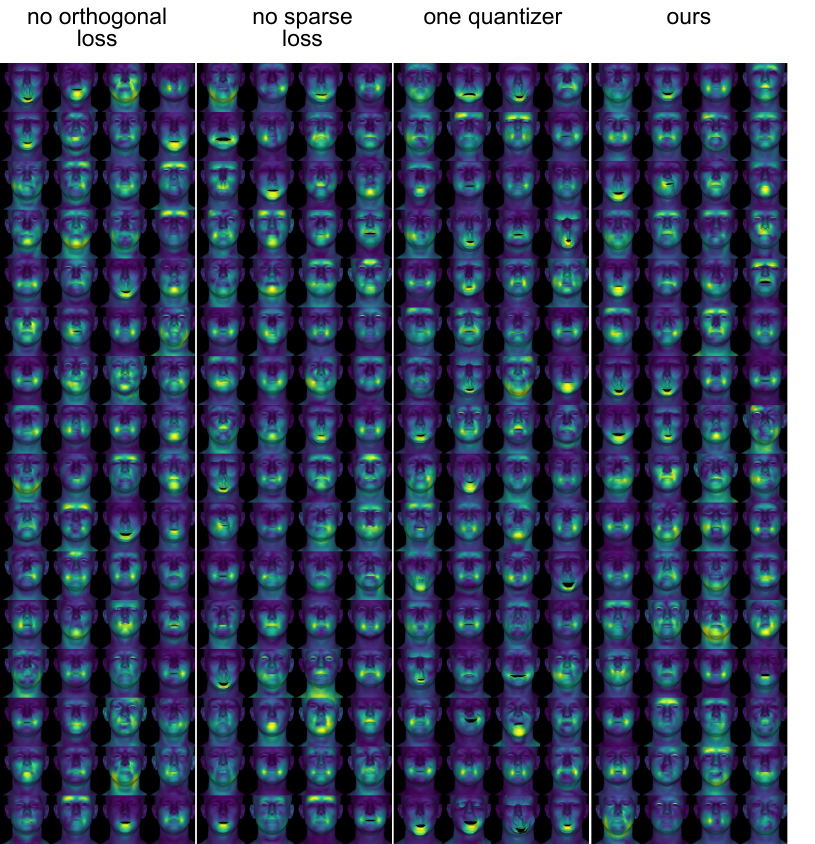}
}
\caption{Impact of design choices on learned codewords. Without orthogonality loss, codewords capture overlapping regions, resulting in redundant templates. Without sparsity loss, regions expand to broad facial areas rather than local discriminative features, reducing interpretability. With only one quantizer, the model fails to capture diverse patterns, yielding a single global template per face instead of a compositional representation.}
\label{fig:interpet_pipeline}
\vspace{-15pt}
\end{figure*}

\section{Ablation Studies}
We conducted an ablation study on the StressID dataset to estimate the impact of model design choices. Results are shown in  Table~\ref{tab:ablation_results}. Decreasing the number of quantizers has the largest effect: reducing from four to a single quantizer, effectively turning it into a VQ-VAE, results in a significant drop in performance. Codebook size also influences performance—both overly small (16) and overly large (256) codebooks result in lower performance. We chose a simple decoder, as the focus of this approach is on the encoder. To demonstrate this, we trained the same model with a more complex and deeper decoder (a 6-layer transformer with a hidden dimension of 128 and 4 attention heads). The larger decoder does not result in improved downstream performance, demonstrating the adequacy of a simple decoder in enabling the training of our encoder. 
Removing the orthogonality or L1 loss leads to improved performance on the binary Stress ID task but reduced performance on the multi-class Stress ID task; in addition, this trade-off is associated with decreased model interpretability.

\begin{figure*}[ht]
\centering
\scalebox{1.0}{
\includegraphics[width=0.9\linewidth, clip]{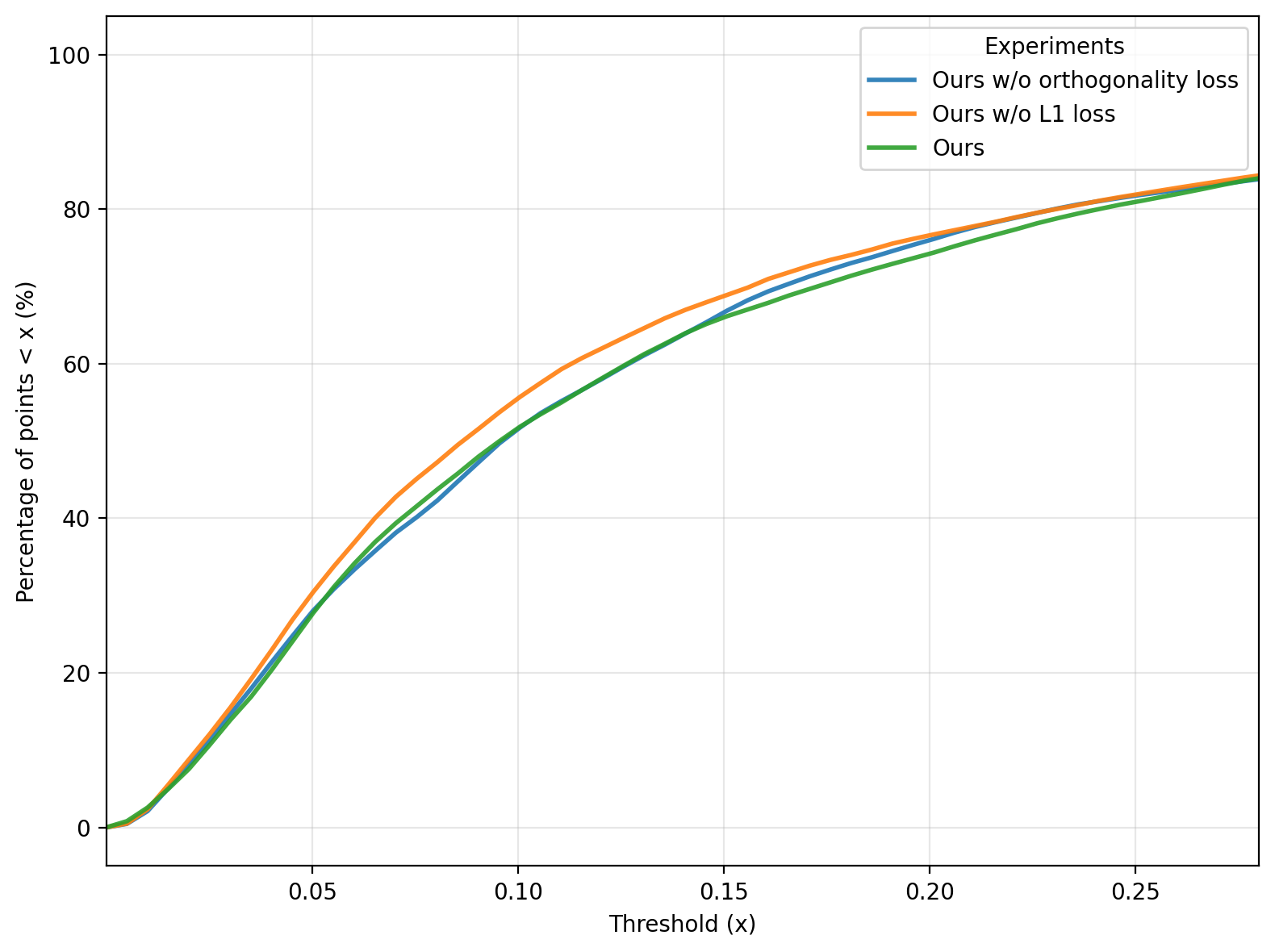}
}
\caption{Percentile curve of displacement points distribution. Shows percentage of points with displacements grater than current value}
\label{fig:sparsity}
\vspace{-15pt}
\end{figure*}

In Figure~\ref{fig:interpet_pipeline} we show that these design choices also affect the learned codebook representations. Without orthogonality loss, the regions of interest captured by different codewords largely overlap, leading to redundant templates. Without sparsity loss, the regions cover broad global areas of the face rather than focusing on local discriminative regions, which reduces interpretability. Finally, with only one quantizer, the model fails to capture diverse facial patterns; faces become non-decomposable to a combination of templates, limiting representational capacity to a single global template per face. We also plot the percentile curve representing the distribution of vertex displacements between the learned facial codebook mesh and the neutral mesh (Figure \ref{fig:sparsity}), as detailed in Section 3.3. This visualization illustrates the number of vertices that undergo a given amount of displacement, providing insight into the variability of the rendered vertices in the learned facial templates. Notably, our method consistently yields the lowest curve, indicating that it produces significantly fewer vertices with large displacements compared to the other two ablation settings. This result demonstrates that our rendered mesh is substantially less scattered.

In Table~\ref{tab:dot_cosine_results}, we evaluate the impact of various components on the orthogonality of the learned representations. Specifically, we assess the similarity between the displacement vectors of facial templates associated with each codeword. 
To quantify this, we compute both the dot product and cosine similarity for all pairs of codewords in the codebook, reporting the average values. Higher scores indicate greater similarity (i.e., more redundant templates), whereas lower scores reflect increased diversity among the templates.
Our method achieves the lowest average dot product and cosine similarity, showing lower redundancy and more unique facial templates compared to configuration without sparsity loss.

\begin{figure*}[t]
\centering
\includegraphics[width=0.75\linewidth]{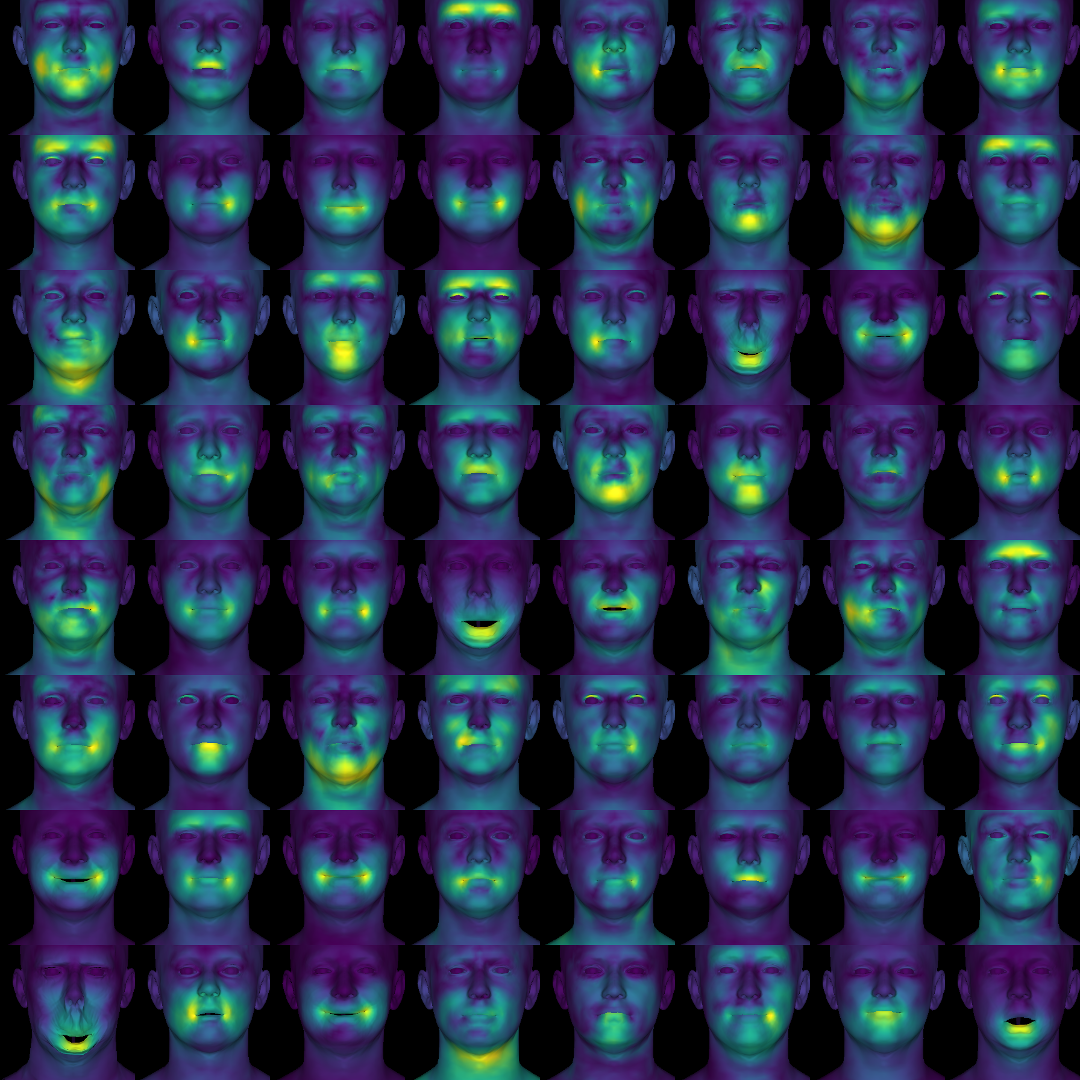}
\caption{Visualization of the learned facial templates.}
\label{fig:all_codebook}
\end{figure*}
\begin{table}[hb]
\centering
\caption{Performance comparison on the StressID dataset. We report F1 Score and Balanced Accuracy for binary and multiclass classification. All values are multiplied by 100 for readability. }

\resizebox{\linewidth}{!}{
\begin{tabular}{lcccc}
\toprule
\multirow{2}{*}{\textbf{Model}} & \multicolumn{2}{c}{\textbf{Binary}} & \multicolumn{2}{c}{\textbf{Multiclass}} \\
 & \textbf{F1} $\uparrow$ & \textbf{BAcc} $\uparrow$ & \textbf{F1} $\uparrow$ & \textbf{BAcc} $\uparrow$ \\
\midrule
Ours w/ Codebook Size 256  & 71.4 & 71.0 & 56.2 & 56.0\\
Ours w/ Codebook Size 16   & 73.7 & 73.3 & 59.6 & 59.0 \\
Ours w/ Single Quantizer (VQ-VAE)   & 70.2 & 69.9 & 52.3 & 51.5 \\
Ours w/ Transformer Decoder & 73.1 & 72.8 & 57.2 & 56.6 \\
Ours w/o orthogonality loss   & 76.0 & 75.7 & 59.4 & 58.8 \\
Ours w/o L1 loss & \textbf{77.4} & \textbf{77.0} & 57.9 & 57.5 \\
Ours               & 73.8 & 73.5 & \textbf{60.3} & \textbf{59.7} \\
\bottomrule
\end{tabular}
}
\label{tab:ablation_results}
\end{table}

\begin{table}[hb]
\centering
\caption{Displacement regions orthogonality comparison on the StressID dataset. We compute dot product and cosine similarity of face displacement vectors corresponding to different codewords. Lower values indicate higher diversity.}

\resizebox{\linewidth}{!}{
\begin{tabular}{lcc}
\toprule
\textbf{Model} & \textbf{Dot product} $\downarrow$ & \textbf{Cosine} $\downarrow$ \\
\midrule
Single Quantizer (VQ-VAE)  & 0.0493 & 0.8676 \\
Ours w/o orthogonality loss     & 0.0158 & 0.8468 \\
Ours w/o L1 loss     & 0.0171 & 0.8390 \\
Ours   & \textbf{0.0086} & \textbf{0.8268} \\
\bottomrule
\end{tabular}
}
\label{tab:dot_cosine_results}
\end{table}

\section{Visualization of all learned facial templates}

We provide a visualization of all learned facial templates in Figure \ref{fig:all_codebook}. Our system successfully captures high-frequency facial movements, including both symmetric and asymmetric motions. This stands in contrast to existing automated Action Unit tools, which are trained on symmetric annotations and thus tend to be biased toward decoding only symmetric facial motions.


\end{document}